\documentclass[runningheads]{llncs}
\usepackage{graphicx}
\usepackage{comment}
\usepackage{amsmath,amssymb} 
\usepackage{color}
\usepackage{multirow}
\usepackage[caption=false]{subfig}
\usepackage{array}
\usepackage{float}
\usepackage{caption}
\usepackage{microtype}
\newcommand{\PreserveBackslash}[1]{\let\temp=\\#1\let\\=\temp}
\newcolumntype{C}[1]{>{\PreserveBackslash\centering}p{#1}}

\begin{document}
\pagestyle{headings}
\mainmatter
\def\ECCVSubNumber{4602}  

\title{Two-Stream Consensus Network for Weakly-Supervised Temporal Action Localization}

\titlerunning{Two-Stream Consensus Network for W-TAL}
%
\author{Yuanhao Zhai\inst{1} \and
Le Wang\inst{1}\thanks{Corresponding author.} \and
Wei Tang\inst{2} \and
Qilin Zhang\inst{3} \and
Junsong Yuan\inst{4} \and
Gang Hua\inst{5}
}
\authorrunning{Y. Zhai, L. Wang, W. Tang, Q. Zhang, J. Yuan, G. Hua}
%
\institute{Xi'an Jiaotong University, Xi'an, Shaanxi, China \and
University of Illinois at Chicago, Chicago, IL, USA \and
HERE Technologies, Chicago, IL, USA \and
State University of New York at Buffalo, Buffalo, NY, USA \and
Wormpex AI Research, Bellevue, WA, USA \\
}
\maketitle

\begin{abstract}
Weakly-supervised Temporal Action Localization (W-TAL) aims to classify and localize all action instances in an untrimmed video under only video-level supervision. 
However, without frame-level annotations, it is challenging for W-TAL methods to identify false positive action proposals and generate action proposals with precise temporal boundaries.
In this paper, we present a Two-Stream Consensus Network (TSCN) to simultaneously address these challenges. 
The proposed TSCN features an iterative refinement training method, where a frame-level pseudo ground truth is iteratively updated, and used to provide frame-level supervision for improved model training and false positive action proposal elimination.
Furthermore, we propose a new attention normalization loss to encourage the predicted attention to act like a binary selection, and promote the precise localization of action instance boundaries.
Experiments conducted on the THUMOS14 and ActivityNet datasets show that the proposed TSCN outperforms current state-of-the-art methods, and even achieves comparable results with some recent fully-supervised methods.
\end{abstract}

\section{Introduction}
The task of Weakly-supervised Temporal Action Localization (W-TAL) aims at simultaneously localizing and classifying all action instances in a long untrimmed video given only video-level categorical labels in the learning phase. Compared to its fully-supervised counterpart, which requires frame-level annotations of all action instances during training, W-TAL greatly simplifies the procedure of data collection and avoids annotation bias of human annotators, therefore has been widely studied \cite{kumar2017hide,wang2017untrimmednets,shou2018autoloc,nguyen2018weakly,paul2018w,alwassel2019refineloc,liu2019completeness,zhai2019action,liu2019weakly,nguyen2019weakly,narayan20193c,yu2019temporal,lee2020background} in recent years.

\begin{figure}[t]
	\centering
	\includegraphics[width=\linewidth]{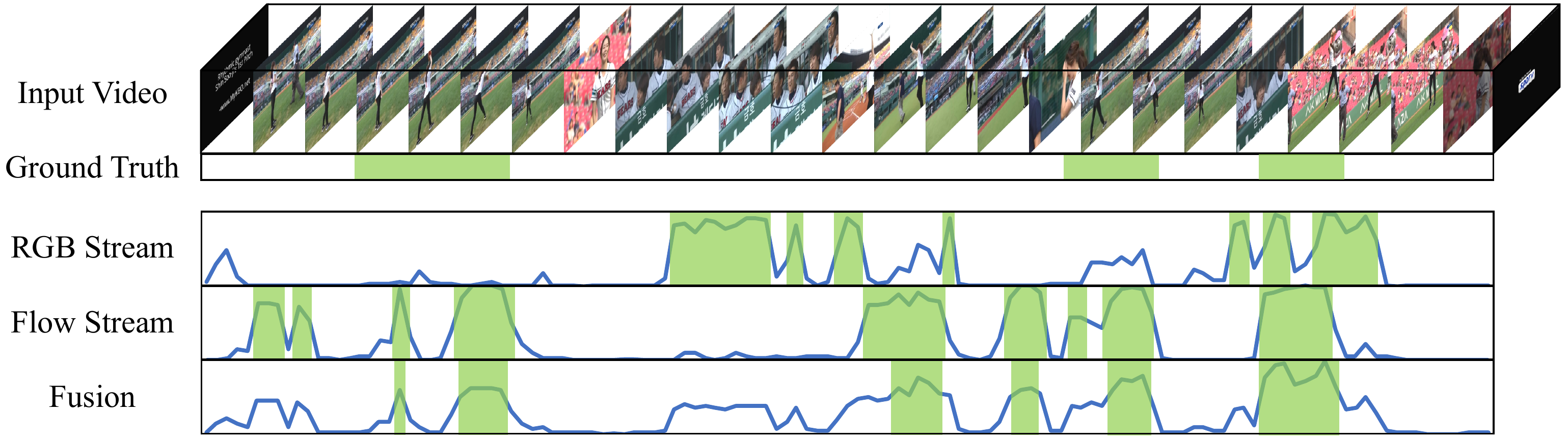}
	\caption{Visualization of two-stream outputs and their late fusion result. The first two rows are an input video and the ground truth action instances, respectively. The last three rows are attention sequences (scaled from $0$ to $1$) predicted by the RGB stream, the flow stream and their weighted sum (\textit{i.e.}, the fusion result), respectively, and the horizontal and vertical axes denote the time and the intensity of attention values, respectively.
	The green boxes denote the localization results generated by thresholding the attention at the value of $0.5$. By properly combining the two different attention distributions predicted by the RGB and flow streams, the late fusion result achieves a higher true positive rate and a lower false positive rate, and thus has better localization performance}
	\label{fig:twoStream}
\end{figure}

Several W-TAL methods \cite{wang2017untrimmednets,paul2018w,nguyen2018weakly,liu2019completeness,nguyen2019weakly,narayan20193c,lee2020background} adopt a Multiple Instance Learning (MIL) framework, where a video is treated as a bag of frames/snippets to perform the video-level action classification. During testing, the trained model slides over time and generates a Temporal-Class Activation Map (T-CAM) \cite{zhou2016learning,nguyen2018weakly} (\textit{i.e.}, a sequence of probability distributions over action classes at each time step) and an attention sequence that measures the relative importance of each snippet. 
The action proposals are generated by thresholding the attention value and/or the T-CAM.
This MIL framework is usually built on two feature modalities, \textit{i.e.,} RGB frames and optical flow, which are fused in two possible ways. \textit{Early fusion} methods \cite{paul2018w,shou2018autoloc,alwassel2019refineloc,liu2019completeness,liu2019weakly,lee2020background} concatenate the RGB and optical flow features before they are fed into the network, and \textit{late fusion} methods \cite{nguyen2018weakly,liu2019completeness,nguyen2019weakly,narayan20193c} compute a weighted sum of their respective outputs before generating action proposals. An example of late fusion is shown in Fig.~\ref{fig:twoStream}.

Despite these recent development, two major challenges still persist. 
One of the most critical problems that prior W-TAL methods suffer from is the lack of ability to rule out false positive action proposals. 
Without frame-level annotations, they localize action instances that do not necessarily correspond to the video-level labels.
For example, a model may falsely localize the action ``swimming" by only checking the existence of water in the scene. Therefore, it is necessary to exploit more fine-grained supervision to guide the learning process.
Another problem lies in the generation of action proposals. 
In previous methods, action proposals are generated by thresholding the activation sequence with a fixed threshold, which is preset empirically.
It has a significant impact on the quality of action proposals: a high threshold may result in incomplete action proposals while a low threshold can bring more false positives. But how to get out of this dilemma was rarely studied.

In this paper, we introduce a Two-Stream Consensus Network (TSCN) to address the two aforementioned problems. 
To eliminate false positive action proposals, we design an iterative refinement training scheme, where a frame-level pseudo ground truth is generated from late fusion attention sequence, and serves as a more precise frame-level supervision to iteratively update two-stream models.
Our intuition is simple: late fusion is essentially a voting ensemble of the RGB and flow streams, and if a proper fusion parameter (\textit{i.e.}, the hyperparameter to control the relative importance of two streams) is selected, late fusion can provide more accurate result compared with each individual stream. 
The advantage of combining these two streams has been demonstrated by the Two-Stream Convolutional Networks \cite{simonyan2014two} for action recognition.
As shown in Fig.~\ref{fig:twoStream}, the two streams produce different activation distributions, which lead to different false positives and false negatives. 
However, when they are combined, the false positive action proposals that only exist in one stream can be largely eliminated, and a high activation value occurs only when both streams are confident that an action instance exists.
Since the late fusion result is of higher quality than single stream result, it can in turn serve as a frame-level pseudo ground truth to supervise and refine both streams.
To generate high-quality action proposals, we introduce a new attention normalization loss. It pushes the predicted attention to approach extreme values, \emph{i.e.}, 0 and 1, so as to avoid ambiguity. As a result, simply setting the threshold to 0.5 yields high-quality action proposals.

Formally, given an input video, RGB and optical flow features are first extracted from pre-trained deep networks. 
Then two-stream base models are trained with video-level labels on RGB and optical flow features, respectively, where the attention normalization loss is used to learn the attention distribution.
After obtaining two-stream attention sequences, a frame-level pseudo ground truth is generated based on their weighted sum (\textit{i.e.}, the late fusion attention sequence), and in turn provides frame-level supervision to improve the two-stream models.
We iteratively update the pseudo ground truth and refine the two-stream base models, and the normalization term at the same time forces the predicted attention to approach a binary selection.
The final localization result is obtained by thresholding the late fusion attention sequence.

To summarize, our contribution is threefold:
\begin{itemize}
	\item We introduce a Two-Stream Consensus Network (TSCN) for W-TAL. The proposed TSCN uses an iterative refinement training method, where a pseudo ground truth generated from late fusion attention sequence at previous iteration can provide more precise frame-level supervision to current iteration.
	\item We propose an attention normalization loss function, which forces the attention to act like a binary selection, and thus improves the quality of action proposals generated by the thresholding method.
	\item Extensive experiments are conducted on two standard benchmarks (\textit{i.e.}, THUMOS14 and ActivityNet) to demonstrate the effectiveness of the proposed method. Our TSCN significantly outperforms previous state-of-the-art W-TAL methods, and even achieves comparable results to some recent fully-supervised TAL methods.
\end{itemize}

\section{Related Work}

\textbf{Action Recognition.} Traditional methods \cite{laptev2005space,dalal2005histograms,dalal2006human,wang2011action} aim to model spatio-temporal information via hand-crafted features. 
Two-Stream Convolutional Networks \cite{simonyan2014two} use two separate Convolutional Neural Networks (CNNs) to exploit appearance and motion clues from RGB frames and optical flow, respectively, and use a late fusion method to reconcile the two-stream outputs. \cite{feichtenhofer2016convolutional} focuses on studying different ways to fuse the two streams. 
The Inflated 3D ConvNet (I3D) \cite{carreira2017quo} expands the 2D CNNs in two-stream networks to 3D CNNs. 
Several recent methods \cite{zhao2018recognize,crasto2019mars,shou2019dmc,wang2019hallucinating,repflow2019} focus on directly learning motion clues from RGB frames instead of calculating optical flow.

\noindent\textbf{Fully-supervised Temporal Action Localization.} Fully-supervised TAL requires frame-level annotations of all action instances during training. Several large-scale datasets have been created for this task, such as THUMOS \cite{jiang2014thumos,gorban2015thumos}, ActivityNet \cite{caba2015activitynet}, and Charades \cite{sigurdsson2016hollywood}. 
Many methods \cite{shou2017cdc,zhao2017temporal,gao2017cascaded,heilbron2017scc,dai2017temporal,xu2017r,lin2018bsn,chao2018rethinking} adopt a two-stage pipeline, \emph{i.e.}, action proposal generation followed by action classification. 
Several methods \cite{xu2017r,dai2017temporal,gao2017turn,chao2018rethinking} adopt the Faster R-CNN \cite{ren2015faster} framework to TAL.
Most recently, some methods \cite{lin2018bsn,long2019gaussian,lin2019bmn} try to generate action proposals with more flexible durations. 
Zeng \textit{et al.} \cite{zeng2019graph} apply the Graph Convolutional Networks (GCN) \cite{kipf2016semi,tan2015learning} to TAL to exploit proposal-proposal relations.

\noindent\textbf{Weakly-supervised Temporal Action Localization.} W-TAL, which only requires video-level supervision during training, greatly relieves the data annotation efforts, and draws more and more attention from the community recently. Hide-and-Seek \cite{kumar2017hide} randomly hides part of the input video to guide the network to discover other relevant parts.  UntrimmedNet \cite{wang2017untrimmednets} consists of a selection module to select the important snippets and a classification module to perform per snippet classification. Sparse Temporal Pooling Network (STPN) \cite{nguyen2018weakly} improves UntrimmedNet by adding a sparse loss to enforce the sparsity of selected segments. W-TALC \cite{paul2018w} jointly optimizes a co-activity similarity loss and a multiple instance learning loss to train the network. AutoLoc \cite{shou2018autoloc} is one of the first two-stage methods in W-TAL, and it first generates initial action proposals and then regresses the boundaries of the action proposals with an Outer-Inner-Contrastive loss. CleanNet \cite{liu2019weakly} improves AutoLoc by leveraging the temporal contrast in snippet-level action classification predictions.
Liu \textit{et al}. \cite{liu2019completeness} propose a multi-branch network to model different stages of action.
Besides, several methods~\cite{nguyen2019weakly,lee2020background} focus on modeling the background and achieve state-of-the-art performances.

Recently, RefineLoc \cite{alwassel2019refineloc} uses an iterative refinement method to help the model capture a \textit{complete} action instance.
And our method is distinct from RefineLoc in three main aspects. 
(1) We adopt a late fusion framework, while RefineLoc adopts an early fusion framework.
(2) Our pseudo ground truth is generated from two-stream late fusion attention sequences, which provides better localization performance than each single stream, while RefineLoc generates the pseudo ground truth by expanding previous localization results, which might result in coarser and over-complete action proposals.
(3) We introduce a new attention normalization loss to explicitly avoid the ambiguity of attention, while RefineLoc has no explicit constraints on attention values.

\begin{figure}[t]
	\centering
	\includegraphics[width=\linewidth]{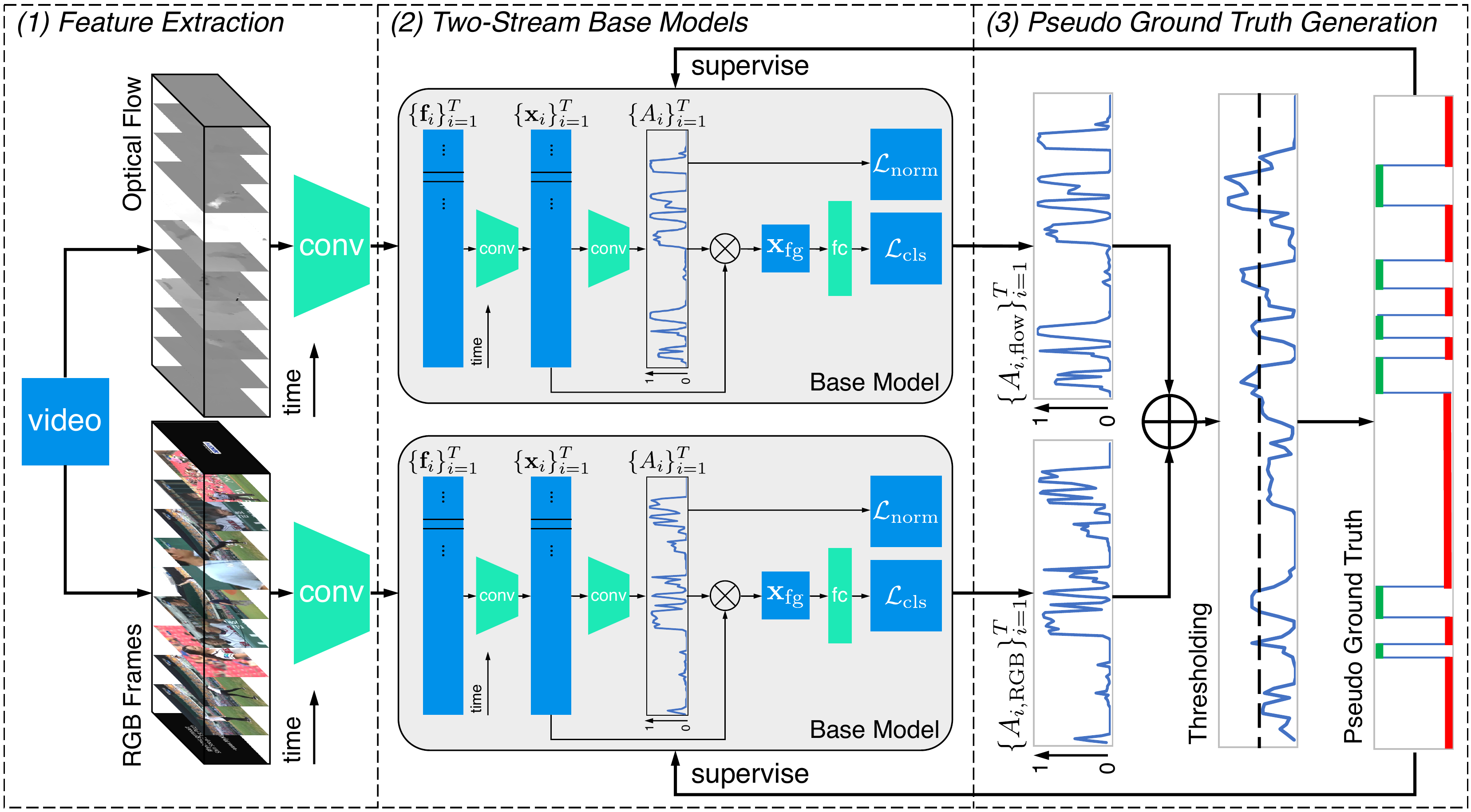}
	\caption{An overview of the proposed Two-Stream Consensus Network, which consists of three parts: (1) RGB and optical flow snippet-level features are extracted with pre-trained models; (2) two-stream base models are separately trained using these RGB and optical flow features; (3) frame-level pseudo ground truth is generated from the two-stream late fusion attention sequence, and in turn provides frame-level supervision to two-stream base models}
	\label{fig:framework}
\end{figure}

\section{Two-Stream Consensus Network}
In this section, we first formulate the task of Weakly-supervised Temporal Action Localization (W-TAL), and then describe the proposed Two-Stream Consensus Network (TSCN) in detail. The overall architecture is shown in Fig.~\ref{fig:framework}.

\subsection{Problem Formulation}
Assume we are given a set of training videos. For each video $v$, we only have its video-level categorical label $\mathbf{y}$, where $\mathbf{y} \in \mathbb{R}^C$ is a normalized multi-hot vector, and $C$ is the number of action categories. The goal of temporal action localization is to generate a set of action proposals $\{ (t_{s}, t_{e}, c, \psi) \}$ for each testing video, where $t_{s}, t_{e}, c, \psi$ denote the start time, the end time, the predicted action category and the confidence score of the action proposal, respectively.

\subsection{Feature Extraction}
Following recent W-TAL methods \cite{shou2018autoloc,nguyen2018weakly,paul2018w,liu2019completeness,liu2019weakly,nguyen2019weakly,narayan20193c,yu2019temporal,lee2020background}, we construct TSCN upon snippet-level feature sequences extracted from the raw video volume. The RGB and optical flow features are extracted with pre-trained deep networks (\textit{e.g.}, I3D \cite{carreira2017quo}) from non-overlapping fixed-length RGB frame snippets and optical flow snippets, respectively. They provide high-level appearance and motion information of the corresponding snippets.
Formally, given a video with $T$ non-overlapping snippets, we denote the RGB features and optical flow features as $\{ \mathbf{f}_{\text{RGB},i} \}_{i=1}^T$ and $\{ \mathbf{f}_{\text{flow},i} \}_{i=1}^T$, respectively, where $\mathbf{f}_{\text{RGB},i}, \mathbf{f}_{\text{flow},i} \in \mathbb{R}^{D}$ are the feature representations of the $i$-th RGB frame and optical flow snippet, respectively, and $D$ denotes the channel dimension.

\subsection{Two-Stream Base Models}
After obtaining the RGB and optical flow features, we first use two-stream base models to perform the video-level action classification, and then iteratively refine the base models with a frame-level pseudo ground truth. The features of two modalities are fed into two separate base models, respectively, and the two base models use the same architecture but do not share parameters. Therefore, in this subsection, we omit the subscript $\text{RGB}$ and $\text{flow}$ for conciseness.

Since the features are not originally trained for the W-TAL task, we concatenate the $T$ input features $\{ \textbf{f}_i \}_{i=1}^{T}$, and use a set of temporal convolutional layers to generate a set of new features $\{ \textbf{x}_i \}_{i=1}^{T}$, where $\textbf{x}_i \in \mathbb{R}^{D'}$, and $D'$ denotes the output feature dimension.

As a video may contain background snippets, to perform video-level classification, we need to select snippets that are likely to contain action instances and meanwhile filter out snippets that are likely to contain background. To this end, an attention value $A_i \in (0, 1)$ to measure the likelihood of the $i$-th snippet containing an action is given by a fully-connected (FC) layer:
\begin{equation}
	A_i = \sigma \left( \mathbf{w}_A \cdot \mathbf{x}_i + b_A \right ),
\end{equation}
where $\sigma (\cdot)$, $\mathbf{w}_A$, and $b_A$ are the sigmoid function, weight vector and bias of the attention layer. We then perform attention-weighted pooling over the feature sequence to generate a single foreground feature $\mathbf{x}_{\text{fg}}$, and feed it to an FC softmax layer to get the video-level prediction:
\begin{equation}
	\mathbf{x}_{\text{fg}} = \frac{1}{\sum_{i=1}^{T} A_{i}} \sum_{i=1}^{T}A_{i} \mathbf{x}_i,
\end{equation}
\begin{equation}
	\hat{y}_{c} = \frac{e^{\mathbf{w}_c \cdot \mathbf{x}_{\text{fg}} + b_c}}{\sum_{i=1}^{C}e^{\mathbf{w}_i \cdot \mathbf{x}_{\text{fg}} + b_i}},
\end{equation}
where $\hat{y}_{c}$ is the probability that the video contains the $c$-th action, and $\mathbf{w}_c$ and $b_c$ are the weight and bias of the FC layer for category $c$. The classification loss function $\mathcal{L}_{\text{cls}}$ is defined as the standard cross entropy loss:
\begin{equation}
	\mathcal{L}_{\text{cls}} = -\sum_{c=1}^{C} y_{c}\log (\hat{y}_{c}),
\end{equation}
where $y_{c}$ denotes the value of label vector $\textbf{y}$ at index $c$.

Ideally, an attention value is expected to be binary, where $1$ indicates the presence of action while $0$ indicates background. Recently, several methods \cite{nguyen2019weakly,lee2020background} introduce a background category, and use the background classification to guide the learning of attention. In this work, instead of using background classification, we introduce an attention normalization term to force the attention to approach extreme values:
\begin{equation}
	\mathcal{L}_{\text{att}} = \frac{1}{l}\min_{\substack{A \subset \{A_i\} \\|A|=l}} \sum_{a \in A}a - \frac{1}{l} \max_{\substack{A \subset \{A_i\} \\|A|=l}} \sum_{a \in A}a,
	\label{eq:lNorm}
\end{equation}
where $l=\max \left(1, \lfloor\frac{T}{s}\rfloor \right)$ and $s$ is a hyperparameter to control the selected snippets. This normalization loss aims to maximize the difference between the average top-$l$ attention values and the average bottom-$l$ attention values, and force the foreground attention to be $1$ and background attention to be $0$. 

Therefore, the overall loss for the base model training is the weighted sum of the classification loss and the attention normalization term:
\begin{equation}
	\mathcal{L}_{\text{base}} = \mathcal{L}_{\text{cls}} + \alpha \mathcal{L}_{\text{att}},
\end{equation}
where $\alpha$ is a hyperparameter to control the weight of the normalization loss.

In addition, the temporal-class activation map (T-CAM) $\{ \textbf{s}_i \}_{i=1}^{T}$, $\textbf{s}_i \in \mathbb{R}^{C}$ is also generated by sliding the classification FC softmax layer over all snippets: 
\begin{equation}
	s_{i,c} = \frac{e^{\mathbf{w}_c \cdot \mathbf{x}_{i} + b_c}}{\sum_{j=1}^{C}e^{\mathbf{w}_j \cdot \mathbf{x}_{i} + b_j}},
\end{equation}
where $s_{i,c}$ is the T-CAM value of $i$-th snippet for category $c$.

\subsection{Pseudo Ground Truth Generation}
We iteratively refine the two-stream base models with a frame-level pseudo ground truth. 
Specifically, we divide the whole training process into several refinement iterations. 
At refinement iteration $0$, only video-level labels are used for training.
And at refinement iteration $n+1$, a frame-level pseudo ground truth is generated at refinement iteration $n$, and provides frame-level supervision for the current refinement iteration. 
However, without \textit{true} ground truth annotations, we can neither measure the quality of the pseudo ground truth, nor guarantee the pseudo ground truth can help the base models achieve higher performance.

Inspired by two-stream late fusion, we introduce a simple yet effective method to generate the pseudo ground truth. Intuitively, locations at which both streams have high activations are likely to contain ground truth action instances; locations at which only one stream has high activations are likely to be either false positive action proposals or true action instances that only one stream can detect; locations at which both streams both have low activations are likely to be the background.

Following this intuition, we use the fusion attention sequence $\{ A_{\text{fuse},i}^{(n)} \}_{i=1}^T$ at refinement iteration $n$ to generate pseudo ground truth $\{ \mathcal{G}_{i}^{(n+1)} \}_{i=1}^T$ for refinement iteration $n + 1$, where $A_{\text{fuse},i}^{(n)}=\beta A_{\text{RGB},i}^{(n)} + (1-\beta)A_{\text{flow},i}^{(n)}$, and $\beta \in [0, 1]$ is a hyperparameter to control the relative importance of RGB and flow attentions.
We introduce two pseudo ground truth generation methods.

\noindent\textbf{Soft pseudo ground truth} means to directly use the fusion attention values as pseudo labels: $\mathcal{G}_{i}^{(n+1)} = A_{\text{fuse},i}^{(n)}$. The soft pseudo labels contain the probability of a snippet being the foreground action, but also add uncertainty to the model.

\noindent\textbf{Hard pseudo ground truth} thresholds the attention sequence to generate a binary sequence:
\begin{equation}
	\mathcal{G}_{i}^{(n+1)} = 
	\left \{
	\begin{array}{lr}
		1, \quad A_{\text{fuse},i}^{(n)} > \theta; & \\
		0, \quad A_{\text{fuse},i}^{(n)} \leq \theta, & \\
	\end{array}
	\right.
\end{equation}
where $\theta$ is the threshold value. Setting a large value of $\theta$ will eliminate the action proposals that only one stream has high activations, and therefore reduces the false positive rate. In contrast, setting a small value of $\theta$ will help models to generate more action proposals and achieve a higher recall. Hard pseudo labels remove the uncertainty and provide stronger supervision, but introduce a hyperparameter.

After generating the frame-level pseudo  ground truth, we force the attention sequence generated by \textit{each} stream to be similar to the pseudo ground truth with a mean square error (MSE) loss\footnote{Although it is straightforward to use a cross entropy loss for hard pseudo ground truth, we found in practice that the cross entropy loss and the MSE loss achieve similar performance. To simplify training, we use the MSE loss for both kinds of pseudo ground truth.}:
\begin{equation}
	\mathcal{L}_{\mathcal{G}}^{(n + 1)} = \frac{1}{T} \sum_{i=1}^{T} \left( A_{i}^{(n + 1)} - \mathcal{G}_{i}^{(n+1)} \right)^2.
\end{equation}
At refinement iteration $n + 1$, the total loss for each stream is 
\begin{equation}
	\mathcal{L}_{\text{total}}^{(n + 1)} = \mathcal{L}_{\text{base}} + \gamma \mathcal{L}_{\mathcal{G}}^{(n + 1)},
\end{equation}
where $\gamma$ is a hyperparameter to control the relative importance of two losses.

\subsection{Action Localization}
During testing, following BaS-Net \cite{lee2020background}, we first temporally upsample the attention sequence and T-CAM by a factor of $8$ via linear interpolation. Then, we select top-$k$ action categories from the fusion video-level prediction $\hat{\textbf{y}}_{\text{fuse}}$ to perform action localization, where $\hat{\textbf{y}}_{\text{fuse}}=\beta \hat{\textbf{y}}_{\text{RGB}} + (1-\beta) \hat{\textbf{y}}_{\text{flow}}$.
For each of these categories, following our intention that the attention performs a binary selection, we generate action proposals by directly thresholding the attention value at $0.5$ and concatenating consecutive snippets. 
The action proposals are scored via a variant of the Outer-Inner-Constrastive score \cite{shou2018autoloc}: instead of using average T-CAM, we use attention weighted T-CAM to measure the outer and inner temporal contrast.
Formally, given action proposal $(t_s, t_e, c)$, fusion attention $\{A_{\text{fuse},i} \}_{i=1}^{T}$ and T-CAM $\{ \textbf{s}_{\text{fuse},i} \}_{i=1}^{T}$, where $\textbf{s}_{\text{fuse},i} = \beta \textbf{s}_{\text{RGB},i} + (1 - \beta)\textbf{s}_{\text{flow},i} $, the score $\psi$ is computed as 
\begin{equation}
	\psi = \frac{\sum_{i=t_s}^{t_e} A_{\text{fuse},i} s_{\text{fuse},i,c}}{t_e - t_s}
	 - \frac{\sum_{i=T_s}^{T_e} A_{\text{fuse},i} s_{\text{fuse},i,c}  - \sum_{i=t_s}^{t_e} A_{\text{fuse},i} s_{\text{fuse},i,c}}{T_e - T_s - (t_e - t_s)},
\end{equation}
where $T_s=t_s-\frac{L}{4}$, $T_e=t_e+\frac{L}{4}$, and $L=t_e-t_s$. We discard action proposals with confidence scores lower than $0$.

\section{Experiments}

\subsection{Dataset and Evaluation}

\textbf{THUMOS14 dataset} \cite{jiang2014thumos} contains $200$ validation videos and $213$ testing videos within $20$ categories for the TAL task. We use the $200$ validation videos for training, and use the $213$ testing videos for evaluation. 

\noindent\textbf{ActivityNet dataset} \cite{caba2015activitynet} has two release versions, \textit{i.e.}, ActivityNet v1.3 and ActivityNet v1.2. ActivityNet v1.3 covers $200$ action categories, with a training set of $10,024$ videos and a validation set of $4,926$ videos.
ActivityNet v1.2 is a subset of ActivityNet v1.3, and covers $100$ action categories, with $4,819$ and $2,383$ videos in the training and validation set, respectively.\footnote{In our experiments, there are $9,937$ and $4,575$ videos in training and validation set of ActivityNet v1.3, respectively, and $4,471$ and $2,211$ videos in training and validation set of ActivityNet v1.2, respectively, because the rest of the videos are unaccessible from YouTube.}
We use the training set and the validation set for training and testing, respectively.

\noindent\textbf{Evaluation Metrics.} Following the standard protocol on temporal action localization, we evaluate our method with mean Average Precision (mAP) under different Intersection-over-Union (IoU) thresholds. We use the evaluation code provided by ActivityNet\footnote{https://github.com/activitynet/ActivityNet/tree/master/Evaluation} to perform the experiments.

\subsection{Implementation Details}
Two off-the-shelf feature extraction backbones are used in our experiments, \textit{i.e.}, UntrimmedNet \cite{wang2017untrimmednets} and I3D \cite{carreira2017quo}, with snippet lengths of $15$ frames and $16$ frames, respectively. 
The two backbones are pre-trained on ImageNet~\cite{Deng2009ImageNet} and Kinetics~\cite{carreira2017quo}, respectively, and are not fine-tuned for fair comparison.
The RGB and flow snippet-level features are extracted at the \texttt{global\_pool} layer as $1024$-D vectors.

The networks are implemented in PyTorch \cite{paszke2017automatic}.
We use the Adam \cite{kingma2014adam} optimizer with a fixed learning rate $0.0001$. 
We train the base models $200$ and $80$ epochs at refinement iteration $0$, and $100$ and $40$ epochs for later refinement iterations for ActivityNet and THUMOS14, respectively.
We set the maximal number of refinement iterations to $4$ for the THUMOS14 dataset, and $24$ for the ActivityNet datasets, and choose base models that achieve the lowest loss at the previous refinement iteration to generate the pseudo ground truth.
To eliminate fragmentary action proposals, temporal max pooling of kernel size $5$ and stride $1$ is used on the fusion attention sequence before pseudo ground truth generation on ActivityNet dataset. 
We use a whole video as a batch.
All hyperparameters are determined via grid search: $s=8$, $\alpha=0.1$, $\beta=0.4$, $\gamma=2$. We set $\theta$ to $0.55$ and $0.5$ for THUMOS14 and ActivityNet, respectively. We choose top-$2$ action categories and also reject categories whose fusion classification prediction scores are lower than $0.1$ to perform action localization.

\begin{table}[t]
	\centering
	\caption{Comparison of our method with state-of-the-art TAL methods on the THUMOS14 testing set. UNT and I3D are abbreviations for UntrimmedNet feature and I3D feature, respectively}
	\label{tab:comparisonOnTHUMOS14}
	\begin{tabular}{c|c|C{0.7cm}C{0.7cm}C{0.7cm}C{0.7cm}C{0.7cm}C{0.7cm}C{0.7cm}C{0.7cm}C{0.7cm}}
	\hline
	\multirow{2}{*}{} & \multirow{2}{*}{Method} & \multicolumn{9}{c}{mAP@IoU (\%)}\\
	& & 0.1 & 0.2 &0.3 & 0.4 & 0.5 & 0.6 & 0.7 & 0.8 & 0.9 \\
	\hline
	\multirow{8}{*}{\rotatebox{90}{Fully-supervised}} & Yuan \textit{et al.} \cite{yuan2017temporal} & 51.0 & 45.2 & 36.5 & 27.8 & 17.8 & - & - & - & - \\
	& CDC \cite{shou2017cdc} & - & - & 40.1 & 29.4 & 23.3 & 13.1 & 7.9 & - & - \\
	& R-C3D \cite{xu2017r} & 54.5 & 51.5 & 44.8 & 35.6 & 28.9 & - & - & - & - \\
	& SSN \cite{zhao2017temporal} & 66.0 & 59.4 & 51.9 & 41.0 & 29.8 & - & - & - & - \\
	& BSN \cite{lin2018bsn} & - & - & 53.5 & 45.0 & 36.9 & 28.4 & 20.0 & - & - \\
	& TAL-Net \cite{chao2018rethinking} & 59.8 & 57.1 & 53.2 & 48.5 & 42.8 & 33.8 & 20.8 & - & - \\
	& GTAN \cite{long2019gaussian} & 69.1 & 63.7 & 57.8 & 47.2 & 38.8 & - & - & - & - \\
	& BMN \cite{lin2019bmn} & - & - & 56.0 & 47.4 & 38.8 & 29.7 & 20.5 & - & - \\
	\hline
	\multirow{16}{*}{\rotatebox{90}{Weakly-supervised}} & UntrimmedNet \cite{wang2017untrimmednets} & 44.4 & 37.7 & 28.2 & 21.1 & 13.7 & - & - & - & - \\
	& STPN (UNT) \cite{nguyen2018weakly} & 45.3 & 38.8 & 31.1 & 23.5 & 16.2 & 9.8 & 5.1 & 2.0 & 0.3 \\
	& AutoLoc (UNT) \cite{shou2018autoloc} & - & - & 35.8 & 29.0 & 21.2 & 13.4 & 5.8 & - & - \\
	& W-TALC (UNT) \cite{paul2018w} & 49.0 & 42.8 & 32.0 & 26.0 & 18.8 & - & 6.2 & - & - \\
	& Liu \textit{et al.} (UNT) \cite{liu2019completeness} & 53.5 & 46.8 & 37.5 & 29.1 & 19.9 & 12.3 & 6.0 & - & - \\
	& RefineLoc (UNT) \cite{alwassel2019refineloc} & - & - & 36.1 & - & 22.6 & - & 5.8 & - & - \\
	& CleanNet (UNT) \cite{liu2019weakly} & - & - & 37.0 & 30.9 & 23.9 & 13.9 & 7.1 & - & - \\
	& BaS-Net (UNT) \cite{lee2020background} & 56.2 & 50.3 & 42.8 & 34.7 & 25.1 & 17.1 & 9.3 & 3.7 & \textbf{0.5} \\ 
	& Ours (UNT) & \textbf{58.9} & \textbf{52.9} & \textbf{45.0} & \textbf{36.6} & \textbf{27.6} & \textbf{18.8} & \textbf{10.2} & \textbf{4.0} & \textbf{0.5} \\
	\cline{2-11}
	& STPN (I3D) \cite{nguyen2018weakly} & 52.0 & 44.7 & 35.5 & 25.8 & 16.9 & 9.9 & 4.3 & 1.2 & 0.1 \\
	& W-TALC (I3D) \cite{paul2018w} & 55.2 & 49.6 & 40.1 & 31.1 & 22.8 & - & 7.6 & - & - \\
	& Liu \textit{et al.} (I3D) \cite{liu2019completeness} & 57.4 & 50.8 & 41.2 & 32.1 & 23.1 & 15.0 & 7.0 & - & - \\
	& RefineLoc (I3D) \cite{alwassel2019refineloc} & - & - & 40.8 & - & 23.1 & - & 5.3 & - & - \\
	& Nguyen \textit{et al.} (I3D) \cite{nguyen2019weakly} & 60.4 & 56.0 & 46.6 & 37.5 & 26.8 & 17.6 & 9.0 & 3.3 & 0.4 \\ 
	& BaS-Net (I3D) \cite{lee2020background} & 58.2 & 52.3 & 44.6 & 36.0 & 27.0 & 18.6 & \textbf{10.4} & \textbf{3.9} & 0.5 \\
	& Ours (I3D) & \textbf{63.4} & \textbf{57.6} & \textbf{47.8} & \textbf{37.7} & \textbf{28.7} & \textbf{19.4} & 10.2 & \textbf{3.9} & \textbf{0.7} \\
	\hline
	\end{tabular} 
\end{table}

\subsection{Comparison with the State-of-the-art}
\textbf{Experiments on THUMOS14}. Table~\ref{tab:comparisonOnTHUMOS14} summarizes the performance comparison between the proposed TSCN and state-of-the-art fully-supervised and weakly-supervised TAL methods on the THUMOS14 testing set. With UntrimmedNet features, TSCN outperforms other W-TAL methods by a large margin, and even achieves comparable results to some recent W-TAL methods with I3D features (\textit{e.g.}, Nguyen \textit{et al}. \cite{nguyen2019weakly} and BaS-Net \cite{lee2020background}) at high IoU thresholds. 

With I3D features, our performance boosts significantly, and outperforms previous W-TAL methods at most IoU thresholds. We note the proposed TSCN can achieve a comparable performance to some recent fully-supervised methods (\textit{e.g.}, R-C3D \cite{xu2017r}). TSCN even outperforms TAL-net \cite{chao2018rethinking} at IoU thresholds $0.1$ and $0.2$. However, as the IoU threshold increases, the performance of TSCN drops significantly, because generating more precise action boundaries need true frame-level ground truth supervision.

\noindent\textbf{Experiments on ActivityNet}. The performance comparisons on ActivityNet v1.2 and v1.3 are shown in Table~\ref{tab:comparsionOnActivityNet1.2} and Table~\ref{tab:comparsionOnActivityNet1.3}, respectively, where our models are trained with I3D features. The proposed TSCN outperforms previous W-TAL methods at the average mAP at IoU threshold $0.5:0.05:0.95$ on both release versions of ActivityNet, verifying the efficacy of our design intuition.

\begin{table}[t]
\hfill
\begin{minipage}[h]{0.48\linewidth}
\centering
\caption{Comparison of our method with state-of-the-art W-TAL methods on the ActivityNet v1.2 validation set. The Avg column indicates the average mAP at IoU thresholds 0.5:0.05:0.95}
\label{tab:comparsionOnActivityNet1.2}
\begin{tabular}{c|C{0.65cm}C{0.65cm}C{0.65cm}|C{0.65cm}} 
	\hline
	\multirow{2}{*}{Method} & \multicolumn{3}{c|}{mAP@IoU (\%)} & \multirow{2}{*}{Avg}\\
	& 0.5 & 0.75 & 0.95 & \\
	\hline
	UntrimmedNet \cite{wang2017untrimmednets} & 7.4 & 3.2 & 0.7 & 3.6 \\
	AutoLoc \cite{shou2018autoloc} & 27.3 & 15.1 & 3.3 & 16.0 \\
	W-TALC \cite{paul2018w} & 37.0 & - & - & 18.0 \\
	Liu \textit{et al.} \cite{liu2019completeness} & 36.8 & 22.0 & 5.6 & 22.4 \\
	Ours & \textbf{37.6} & \textbf{23.7} & \textbf{5.7} & \textbf{23.6} \\
	\hline
\end{tabular}
\end{minipage}
\hfill
\begin{minipage}[h]{0.48\linewidth}
\centering
\caption{Comparison of our method with state-of-the-art W-TAL methods on the ActivityNet v1.3 validation set. The Avg column indicates the average mAP at IoU thresholds 0.5:0.05:0.95}
\label{tab:comparsionOnActivityNet1.3}
\begin{tabular}{c|C{0.65cm}C{0.65cm}C{0.65cm}|C{0.65cm}} 
	\hline
	\multirow{2}{*}{Method} & \multicolumn{3}{c|}{mAP@IoU (\%)} & \multirow{2}{*}{Avg}\\
	& 0.5 & 0.75 & 0.95 & \\
	\hline
	STPN \cite{nguyen2018weakly} & 29.3 & 16.9 & 2.7 & - \\
	Liu \textit{et al.} \cite{liu2019completeness} & 34.0 & 20.9 & \textbf{5.7} & 21.2 \\
	Nguyen \textit{et al.} \cite{nguyen2019weakly} & \textbf{36.4} & 19.2 & 2.9 & - \\
	Ours & 35.3 & \textbf{21.4} & 5.3 & \textbf{21.7} \\
	\hline
\end{tabular}  
\end{minipage}
\hfill 
\end{table}

\subsection{Ablation Study}
\label{subsec:ablationStudy}

In this subsection, a set of ablation studies is conducted on the THUMOS14 testing set with UntrimmedNet feature to analyze the efficacy of each component in the proposed TSCN.

\begin{figure}[t]
	\begin{minipage}[h]{0.5\linewidth}
		\centering
		\captionof{table}{Comparison of our method with different attention normalization functions on the THUMOS14 testing set. $\mathcal{L}_{bg}$ is the background classification loss introduced in \cite{nguyen2019weakly}, and $\mathcal{L}_{att}$ is defined in Equation~\eqref{eq:lNorm}. The var column denotes the average attention variance over the whole testing set}
		\label{tab:comparisonWithDifferentNormalization}
		\begin{tabular}{ccc|C{0.8cm}C{0.8cm}C{0.8cm}|c} 
			\hline
			\multirow{2}{*}{$\mathcal{L}_{\text{cls}}$} & \multirow{2}{*}{$\mathcal{L}_{bg}$} & \multirow{2}{*}{$\mathcal{L}_{att}$} & \multicolumn{3}{c|}{mAP@IoU (\%)} & \multirow{2}{*}{Var} \\
			& & & 0.3 & 0.5 & 0.7 & \\
			\hline
			\checkmark & - & - & 29.6 & 16.1 & 4.1 & 0.0440  \\
			\checkmark & \checkmark & - & 34.3 & 19.3& 6.7 & 0.0599 \\ 
			\checkmark & - & \checkmark & \textbf{40.9} & \textbf{24.0} & \textbf{8.2} & \textbf{0.0937} \\ 
			\checkmark & \checkmark & \checkmark & 40.6 & 23.6 & 7.8 & 0.0886 \\ 
			\hline
		\end{tabular}
	\end{minipage}
	\begin{minipage}[h]{0.5\linewidth}
		\centering
		\includegraphics[width=0.83\linewidth]{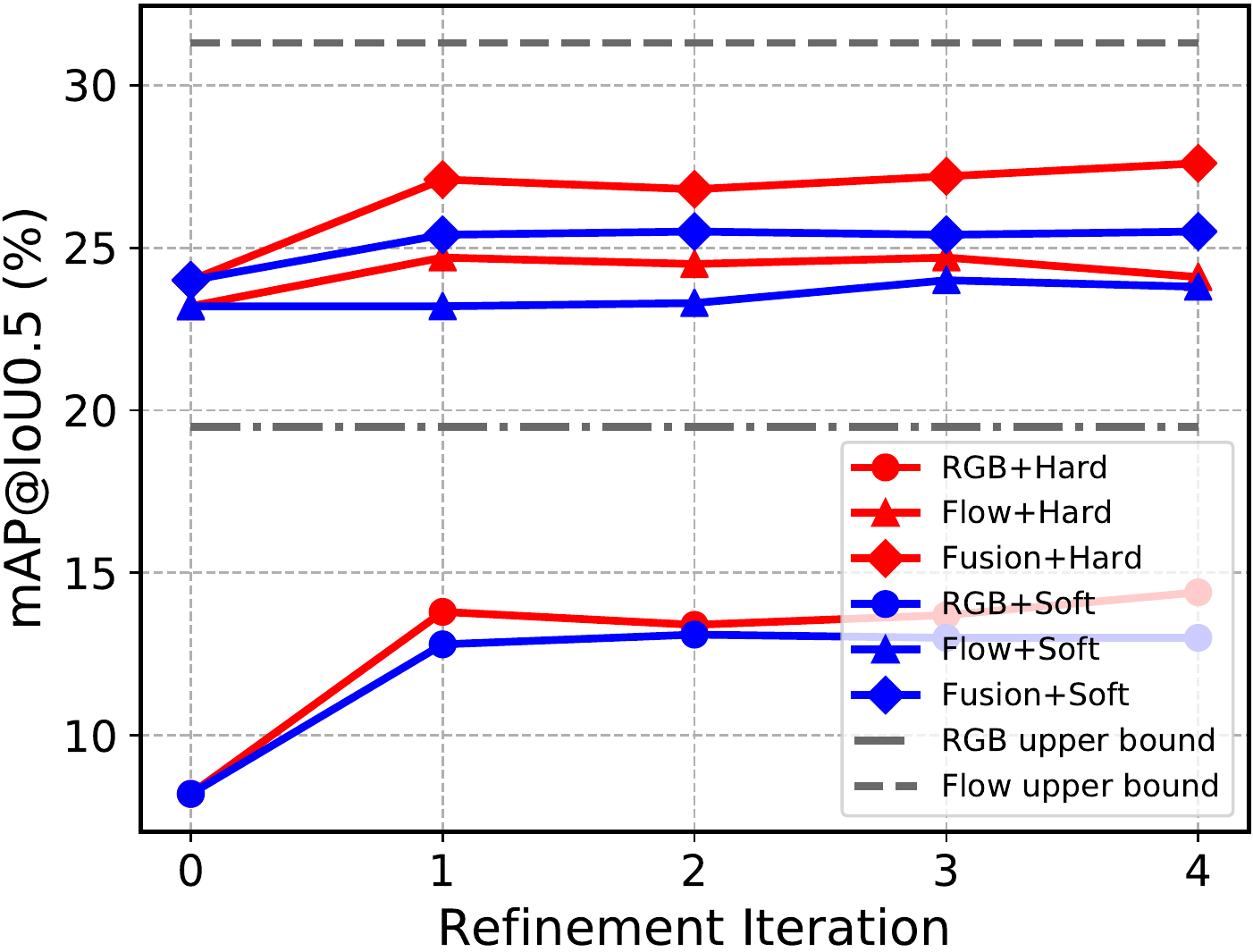}
		\caption{Comparison between models trained with different pseudo ground truth on the THUMOS14 testing set. The upper bounds denote models trained with ground truth actionness sequence}
		\label{fig:comparisonWithDifferentPseudoGT}
	\end{minipage}
\end{figure}

\noindent\textbf{Ablation study on $\mathcal{L}_{\text{att}}$}. 
The goal of $\mathcal{L}_{\text{att}}$ in Equation~\eqref{eq:lNorm} is to force the attention values to approach extreme values, and therefore generate a clean foreground feature $\textbf{x}_{\text{fg}}$ and improve action proposal quality.
Some recent methods~\cite{nguyen2019weakly,lee2020background} introduce background classification to W-TAL. 
Particularly, background classification loss $\mathcal{L}_{bg}$~\cite{nguyen2019weakly} is introduced to classify the background, where a background attention is defined as $1-A_i$, and a background feature is generated via background attention-weighted pooling over all snippets to perform the background classification.
Therefore, $\mathcal{L}_{bg}$ is in essence an implicit attention normalization loss.
However, one drawback of such background loss is that assigning background labels to all videos will make the value of the background category in the T-CAM increase.
We reproduce $\mathcal{L}_{bg}$ in our model, compare it with our proposed $\mathcal{L}_{\text{att}}$, and list the results in Table~\ref{tab:comparisonWithDifferentNormalization}. 
The results reveal that both $\mathcal{L}_{bg}$ and $\mathcal{L}_{\text{att}}$ help improve the performance. 
And the proposed $\mathcal{L}_{\text{att}}$ achieves higher attention variance and better localization performance than $\mathcal{L}_{bg}$, demonstrating that the our attention normalization term $\mathcal{L}_{\text{att}}$ can better avoid the ambiguity of attention.
Surprisingly, with both $\mathcal{L}_{bg}$ and $\mathcal{L}_{\text{att}}$, the localization performance is still lower than that with only $\mathcal{L}_{\text{att}}$, and we think this is because the noise of background classification reduces the accuracy of action proposal scores.

\noindent\textbf{Ablation study on Pseudo Ground Truth}.
Fig.~\ref{fig:comparisonWithDifferentPseudoGT} plots performance comparison between different pseudo ground truth methods at different refinement iterations. 
Both soft and hard pseudo ground truth help improve the localization performance. 
The hard pseudo ground truth removes uncertainty to the model, and thus achieves higher performance improvement.
However, with the same frame-level supervision, the flow stream outperforms the RGB stream by a large margin. We think this is because of the nature of two modalities: the RGB modality is less sensitive to actions than the optical flow modality. 
To demonstrate this, we generate a \textit{true} frame-level ground truth actionness sequence (action categories are not used), train our model in the same way as the pseudo ground truth. The results are plotted in Fig.~\ref{fig:comparisonWithDifferentPseudoGT} as an upper bound. The results verify our hypothesis and demonstrate that the optical flow modality is more suitable for the action localization task than the RGB modality. 

\begin{table}[t]
	\centering
	\caption{Comparison between the model trained with only video-level labels and the model trained with hard pseudo ground truth on the THUMOS14 testing set. The label column denotes the supervision used in training, where ``video" indicates only video-level labels are leveraged, and ``frame" indicates the hard pseudo ground truth is also leveraged during training. Precision, recall and F-measure are calculated under IoU threshold $0.5$}
	\label{tab:}
	\begin{tabular}{c|C{1cm}|C{0.7cm}C{0.7cm}C{0.7cm}C{0.7cm}C{0.7cm}|c|c|c}
	\hline
	\multirow{2}{*}{Modality} & \multirow{2}{*}{Label} & \multicolumn{5}{c|}{mAP@IoU (\%)} & \multirow{2}{*}{Precision (\%)} & \multirow{2}{*}{Recall (\%)} & \multirow{2}{*}{F-measure} \\
	& & 0.3 & 0.4 & 0.5 & 0.6 & 0.7 & & &  \\
	\hline
	RGB & video & 19.8 & 13.2 & 8.2 & 4.5 & 1.9 & 10.2 & 20.9 & 0.1371 \\
	RGB & frame & 31.4 & 22.1 & 14.4 & 8.9 & 5.2 & 20.9 & 30.8 & 0.2489 \\
	\hline
	Flow & video & 40.2 & 32.0 & 23.2 & 15.4 & 7.2 & 25.5 & 43.3 & 0.3207 \\
	Flow & frame & 40.8 & 32.7 & 24.1 & 16.8 & 8.7 & 30.9 & 42.4 & 0.3573 \\
	\hline
	Fusion & video & 40.9 & 32.4 & 24.0 & 15.9 & 8.2 & 23.6 & 44.4 & 0.3078 \\
	Fusion & frame & 45.0 & 36.5 & 27.6 & 18.8 & 10.2 & 31.3 & 44.6 & 0.3680 \\
	\hline
	\end{tabular} 
\end{table}

Table~\ref{eq:lNorm} lists the detailed performance comparison between the model trained with only video-level labels and that trained with the hard pseudo ground truth. The results show that pseudo ground truth improves the localization performance for both modalities at all IoU thresholds, and thus improves the performance of the fusion result. Also, the pseudo ground truth greatly improves the precision and recall for the RGB stream and the fusion result, and improves the precision for the flow stream with a minor loss of recall (the overall F-measure improves significantly), which demonstrates that the pseudo ground truth can help eliminate false positive action proposals.

\begin{figure}[t]
	\centering
	\includegraphics[width=0.99\linewidth]{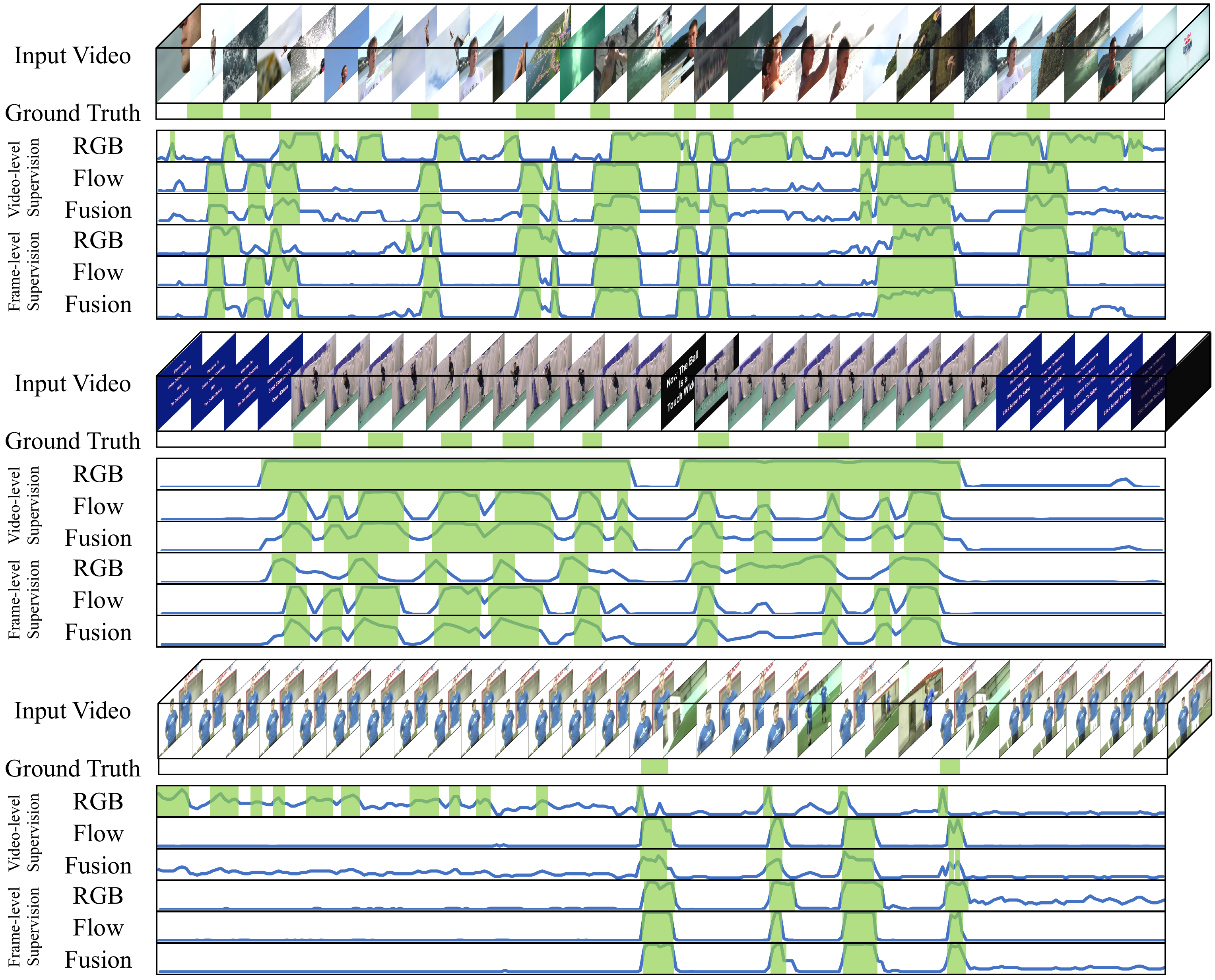}
	\caption{Qualitative results on the THUMOS14 testing set. The eight rows in each example are input video, ground truth action instance, RGB stream, flow stream, and fusion attention sequences from the model trained with only video-level labels and frame-level pseudo ground truth, respectively. Action proposals are represented by green boxes. The horizontal and vertical axes are time and intensity of attention, respectively}
	\label{fig:qualitativeResult}
\end{figure}

\noindent\textbf{Qualitative Analysis.} Three representative examples of TAL results are plotted in Fig.~\ref{fig:qualitativeResult} to illustrate the efficacy of the proposed pseudo supervision. 
In the first example of diving and cliff diving, with only video-level labels, the RGB stream provides worse localization result than the flow stream, and thus leads to a noisy fusion attention sequence. The pseudo ground truth guides the RGB stream to identify false positive action proposals and discover true action instances, and further leads to a cleaner fusion attention sequence, where high activations correspond better to the ground truth.
In the second example of cricket shot, with only video-level supervision, the RGB stream can only distinguish certain scenes, and fails to separate proximate action instances. In contrast, the flow stream can precisely detect action instances. Therefore, the pseudo ground truth helps the RGB stream to separate consecutive action instances.
In the last example of soccer penalty, both streams have high activations on certain false positive temporal locations. Under this circumstance, the false positive action proposals will have higher activations under frame-level pseudo supervision. To eliminate such false positive action proposals, however, need true ground truth supervision.
To summarize, the two modalities have their own strengths and limitations: the RGB stream is sensitive to appearance, thus it fails in scenes shot from unusual angles or separating proximate action instances in the same scene; the flow stream is sensitive to motion, and provides more accurate results, but it fails in slow or occluded motion.
Qualitative results reveal that the pseudo ground truth helps two streams reach a consensus at most temporal locations. 
Therefore, the fusion attention sequence becomes cleaner and helps generate more precise action proposals and more reliable confidence scores.

\section{Conclusions}
In this paper, we propose a Two-Stream Consensus Network (TSCN) for W-TAL, which benefits from an iterative refinement training method and a new attention normalization loss. 
The iterative refinement training uses a novel frame-level pseudo ground truth as fine-grained supervision, and iteratively improves the two-stream base models.
The attention normalization loss function reduces the ambiguity of attention values, and thus leads to more precise action proposals.
Experiments on two benchmarks demonstrate the proposed TSCN outperforms current state-of-the-art methods, and verify our design intuition.

\section*{Acknowledgement}
This work was supported partly by National Key R\&D Program of China Grant 2018AAA0101400, NSFC Grants 61629301, 61773312, and 61976171, China Postdoctoral Science Foundation Grant 2019M653642, Young Elite Scientists Sponsorship Program by CAST Grant 2018QNRC001, and Natural Science Foundation of Shaanxi Grant 2020JQ-069.

\clearpage
%
%

\bibliographystyle{splncs04}
\bibliography{egbib}
\end{document}